\pdfoutput=1

\documentclass[11pt]{article}

\usepackage{authblk}
\usepackage[]{acl}

\usepackage{times}
\usepackage{latexsym}

\usepackage[T1]{fontenc}

\usepackage[utf8]{inputenc}

\usepackage{microtype}

\usepackage{inconsolata}

\usepackage{graphicx}
\usepackage{booktabs}
\usepackage{multirow} 
\usepackage{makecell}
\usepackage{subcaption}
\usepackage{amsmath}
\usepackage{array}
\usepackage{longtable}

\usepackage{CJKutf8}

\newcommand*{\affaddr}[1]{#1} 
\newcommand*{\affmark}[1][*]{\textsuperscript{#1}}
\newcommand*{\email}[1]{\texttt{#1}}

\title{Learning to Trust Your Feelings:\\Leveraging Self-awareness in LLMs for Hallucination Mitigation}

\author{%
Yuxin Liang\thanks{Work done in IDEA}\affmark[1], Zhuoyang Song\affmark[2], Hao Wang\affmark[1], Jiaxing Zhang\affmark[2]\\
\affaddr{\affmark[1]X2Robot} \\
\affaddr{\affmark[2]International Digital Economy Academy}\\
\email{liangyuxin42@gmail.com}, \email{wanghao@x2robot.com}\\
\email{\{songzhuoyang, zhangjiaxing\}@idea.edu.cn}\\
}

\begin{document}

\maketitle

\begin{abstract}


We evaluate the ability of Large Language Models (LLMs) to discern and express their internal knowledge state, a key factor in countering factual hallucination and ensuring reliable application of LLMs.
We observe a robust self-awareness of internal knowledge state in LLMs, evidenced by over 85\% accuracy in knowledge probing.
However, LLMs often fail to express their internal knowledge during generation, leading to factual hallucinations. 
We develop an automated hallucination annotation tool, DreamCatcher, which merges knowledge probing and consistency checking methods to rank factual preference data.
Using knowledge preference as reward, We propose a Reinforcement Learning from Knowledge Feedback (RLKF) training framework, leveraging reinforcement learning to enhance the factuality and honesty of LLMs. 
Our experiments across multiple models show that RLKF training effectively enhances the ability of models to utilize their internal knowledge state, boosting performance in a variety of knowledge-based and honesty-related tasks.


\end{abstract}

\section{Introduction}
\label{sec:Introduction}

\begin{figure}
    \centering
    \includegraphics[width=\linewidth]{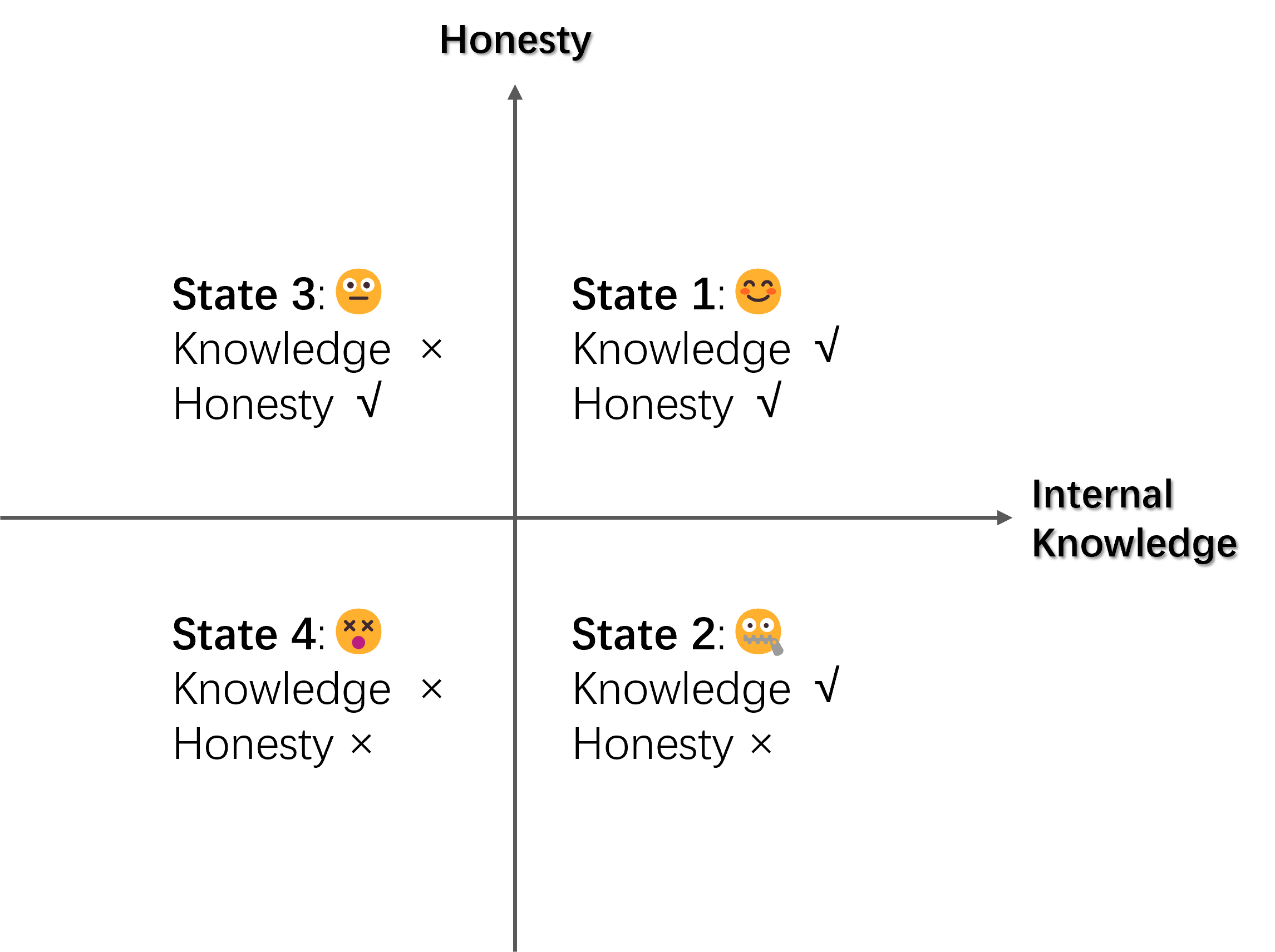}
    \caption{Internal knowledge state categorization of LLM, based on the possession of corresponding internal knowledge and the capacity to express it honestly.}
    \label{fig:hallucination-categorize}
\end{figure}

Large Language Models (LLMs), including notable examples such as GPT-3 \cite{brown2020language}, LLaMA (\citet{touvron2023llama1}, \citet{touvron2023llama2}), and PaLM \cite{chowdhery2023palm}, have emerged as a transformative tool in diverse fields due to their robust capabilities in various tasks. However, despite this significant progress and success, an inherent challenge continues to persist: their tendency to "hallucinate", i.e., generate content misaligned with actual facts. This issue is particularly problematic in critical applications, such as clinical or legal scenarios, where the generation of reliable and accurate text is vital. Therefore, mitigating hallucinations in LLMs is a crucial step toward enhancing their practical application scope and improving the overall trust in these emerging technologies.

Hallucinations of LLMs can be categorized into three types \cite{zhang2023siren}: input conflict, context conflict, and factual conflict. 
This paper focus on the issue of fact-conflicting hallucination, where LLM produces fluent and seemingly plausible content, but conflicts with real-world facts, pose risks of misleading users and compromise the models' fact-based reasoning.

Commonly used hallucination mitigation methods, such as retrieval augmentation generation (RAG), address fact-conflict hallucination of LLM by bringing in external knowledge, but at the cost of introducing a costly and complex retrieval system. In this paper, we propose to mitigate the factual hallucination problem from the perspective of enhancing the model's utilization of internal knowledge.

Previous works (\citet{kadavath2022language}, \citet{azaria2023internal}, \citet{agrawal2023language}) have shown that LLMs have the capability to discern the validity of factual statements, supported further by \citet{kadavath2022language} suggesting these models' capacity to assess their ability in responding to specific questions. Nevertheless, the universality and extent of models' self-awareness of their knowledge remains an open question.
In light of this, we conducted exploratory experiments to probe the knowledge state of various models across different scales, employing linear probes to ascertain the accuracy of models' judgments regarding their internal knowledge states. The results revealed that all models under analysis demonstrated proficient accuracy in recognizing whether they have the internal knowledge about certain facts.

However, during generation, such accurate judgments do not translate into honest output; instead, in the absence of specific internal knowledge, models often manifest a tendency towards hallucinations. Therefore, to mitigate factual hallucinations, it is crucial that models leverage their self-assessed judgments concerning their knowledge status.

We propose a training framework named reinforcement learning from knowledge feedback (RLKF) to improve the factuality and honesty of LLM with reinforcement learning using factual preferences as reward. 
Through the hallucination annotation method DreamCatcher – a blend of knowledge probing and consistency-based judgments – we rank the knowledge-based Question-Answering (QA) data adhering to a preference hierarchy delineated as: $factuality > uncertainty > hallucination$. This factual preference data is then utilised to train the reward model which is deployed to optimize the Large Language Model via Proximal Policy Optimisation (PPO) algorithm.


The primary contributions of this paper are articulated as follows:
\begin{enumerate}
    \item We carried out extensive experiments on different models' capacity to discern their own internal knowledge. The results indicate that LLMs are highly adept at discerning their internal knowledge, with an impressive accuracy over \textbf{85\%} in most cases with a limited amount of data.
    
    \item We develop \textbf{DreamCatcher\footnote{https://github.com/liangyuxin42/dreamcatcher}}, an automatic hallucination detection tool for scoring the degree of hallucination in LLM generations. DreamCatcher integrates knowledge probing methods and consistency judgments, achieving 81\% agreement with human annotator.
    
    \item We introduce the Reinforcement Learning from Knowledge Feedback (RLKF) training framework to optimize LLM against the factual preference. The experiment results on multiple knowledge and reasoning tasks indicate that RLKF not only enhances the honesty and factuality of LLMs but also improves their general capabilities.
\end{enumerate}


\section{Problem Setup}
\label{sec:problem setup}

\begin{figure*}[htbp]
\begin{subfigure}{0.33\textwidth}
\includegraphics[width=0.95\linewidth]{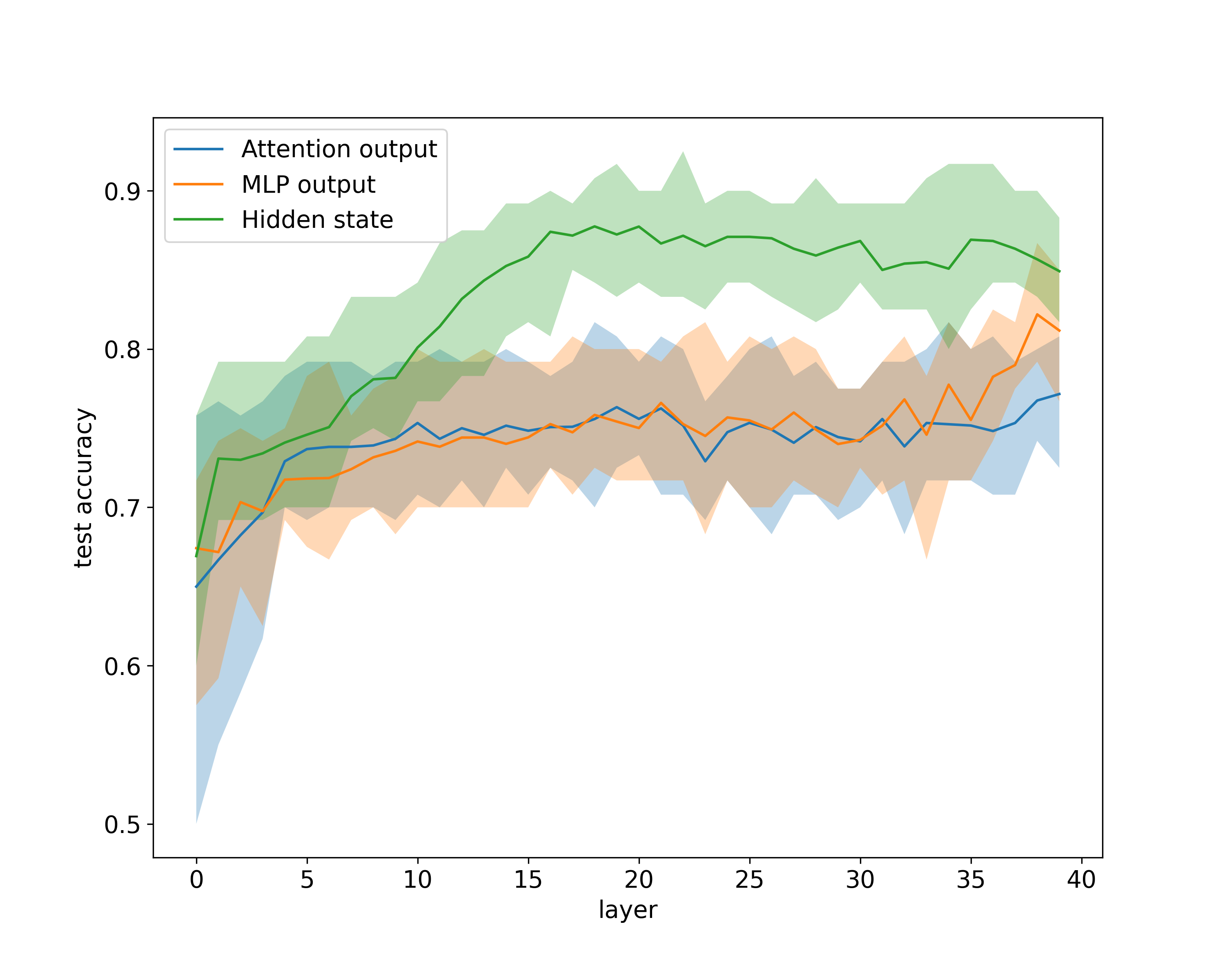} 
\caption{Llama2-chat-13B}
\label{fig:subim1}
\end{subfigure}
\begin{subfigure}{0.33\textwidth}
\includegraphics[width=0.95\linewidth]{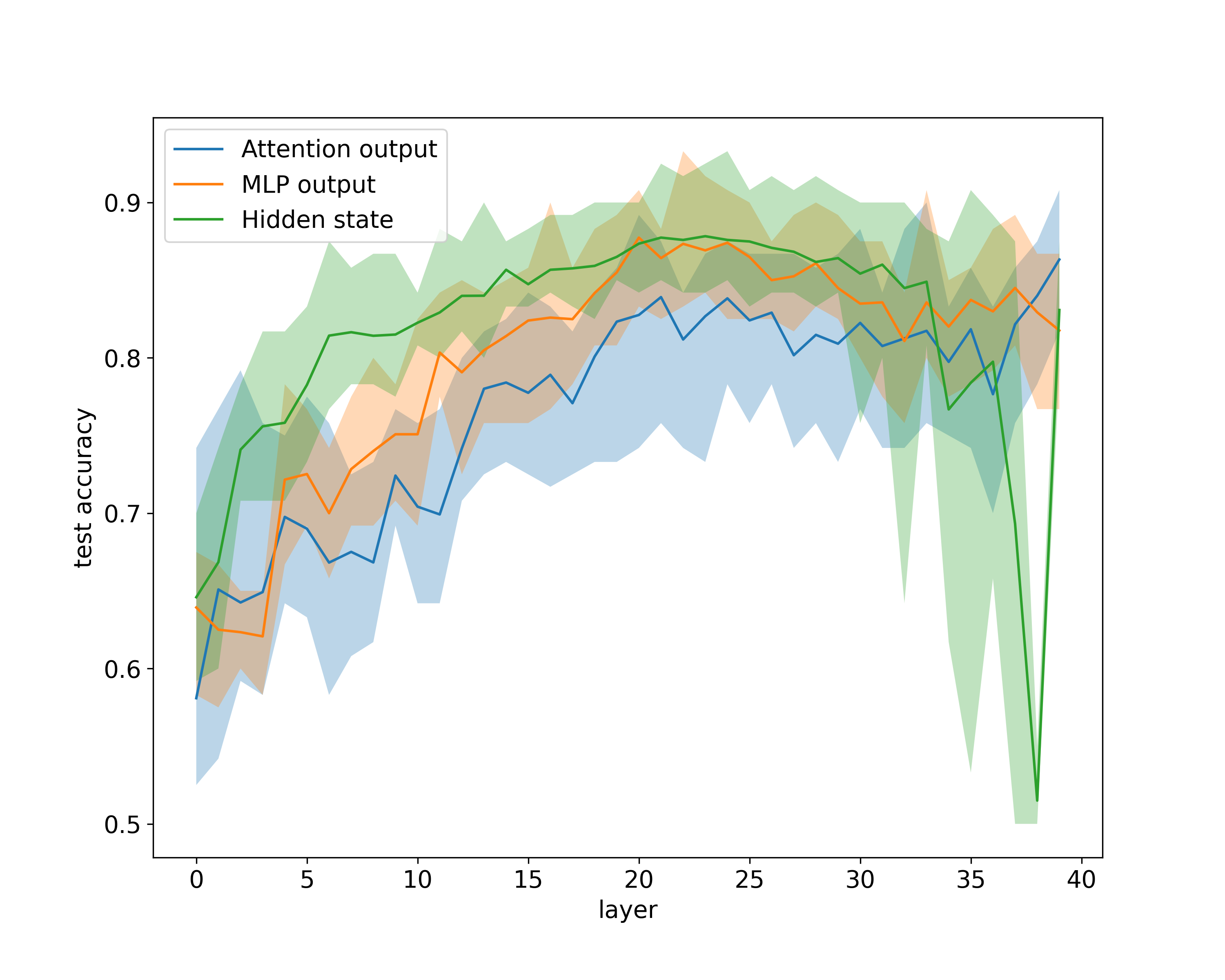}
\caption{Qwen-chat-14B}
\label{fig:subim2}
\end{subfigure}
\begin{subfigure}{0.33\textwidth}
\includegraphics[width=0.95\linewidth]{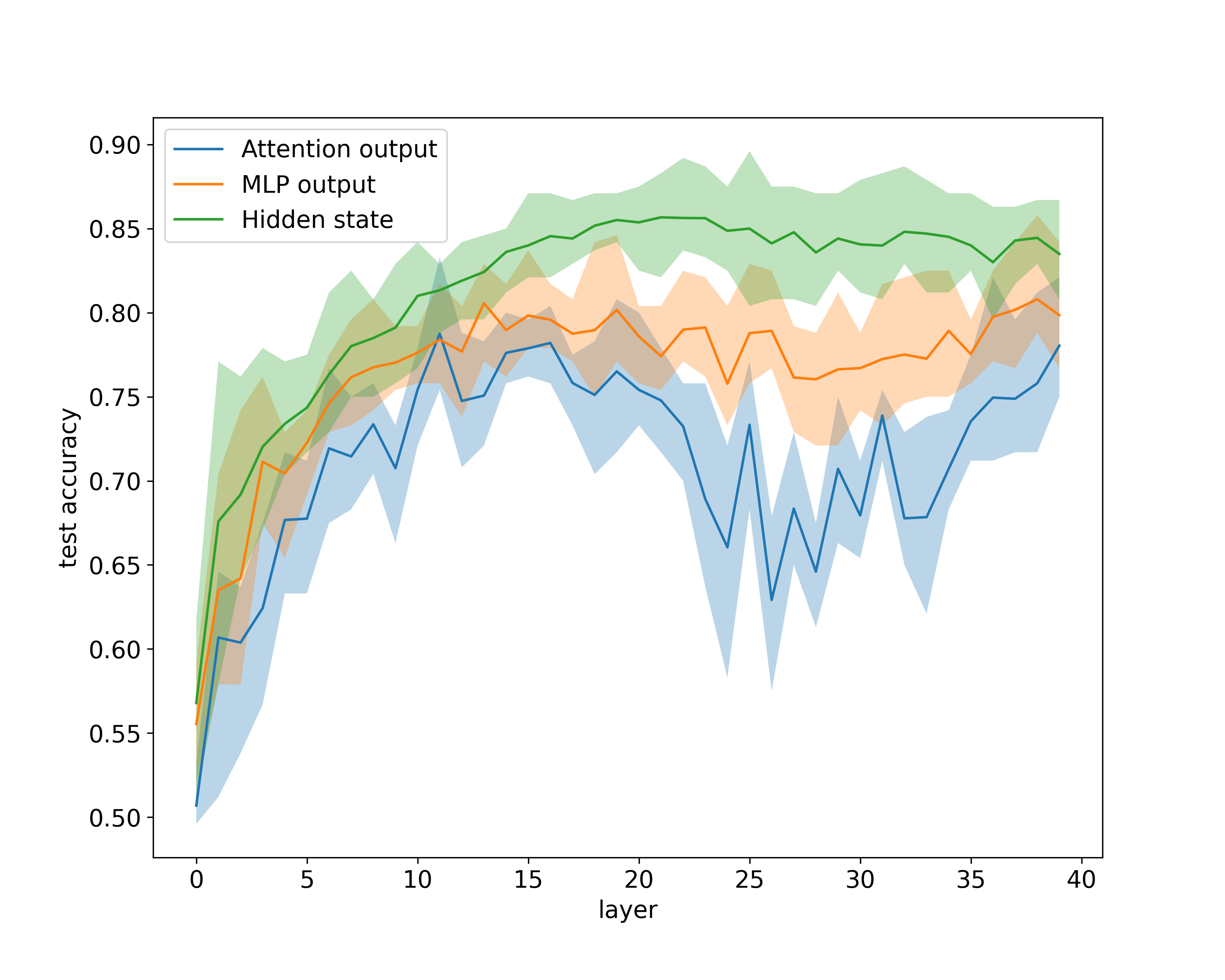}
\caption{Ziya-reader-13B}
\label{fig:subim3}
\end{subfigure}
\caption{Accuracy of knowledge state probing across different models with different internal representations. The light-colored area in the figure shows the range of accuracy for ten repetitions of the experiment, and the solid line shows the mean accuracy. More results shown in \ref{sec:More probing results}}
\label{fig:probe-results}
\end{figure*}


Hallucination, in the context of Large Language Models, refers to a set of inconsistencies during model generation. The central focus of this paper is exploring the fact-conflict hallucination which is defined as the inconsistency between the generated content of the model and the established facts. Although the definition provides a description of the generation results, the causes underlying this phenomenon are multifaceted.

In general, LLMs encode factual knowledge into parameters during training and utilize this internal knowledge for generation during inference. However, LLMs do not always honestly express the knowledge in its parameters, which is one of the major causes of fact-conflict hallucination.

For a given question that requires factual knowledge, the model output can be classified into one of four states, depending on the model's internal knowledge and its honesty. These states are illustrated in Figure \ref{fig:hallucination-categorize}:

\textbf{State 1}: The model has relevant internal knowledge and expresses it faithfully.

\textbf{State 2}: Despite having the relevant internal knowledge, the model fails to express it honestly. This discrepancy could be due to various factors such as the decoding strategy \citep{lee2022factuality, chuang2023dola}, hallucination snowballing \citep{zhang2023language}, or misalignment issues \citep{schulmanreinforcement}.

\textbf{State 3}: The model lacks the necessary internal knowledge but honestly indicates unawareness.

\textbf{State 4}: The model lacks the necessary internal knowledge and instead produces a hallucinated response.

Outputs in \textbf{State 2} and \textbf{State 4} are both considered forms of hallucination, despite the differing conditions of internal knowledge.

In the upper section of Figure \ref{fig:hallucination-categorize}, the model's outputs are devoid of hallucinations, honestly mirroring its internal knowledge reservoir. Here, \textbf{State 1} stands out as the most desirable state, where the model both possesses and faithfully produces the relevant knowledge.

Many efforts have been deployed to transition model toward state 1. 

Retrieval-augmented generation (RAG) attempts to bypass the lack of internal knowledge by providing knowledge via context, thereby enabling the model to transition from \textbf{State 3/4} to \textbf{State 1}. 
On another front, certain strategies, like those of \citet{li2023inference} and \citet{chuang2023dola}, seek to move the model from \textbf{State 2} to \textbf{State 1} by intervening the model's internal representation or the decoding process during inference. 
While these methods improve the model's capacity to express existing internal knowledge, they disregard scenarios where the model lacks relevant internal knowledge. Also, interference at inference time can potentially lead to unpredictable effects on other types of tasks.

Without the introduction of external knowledge, the mitigation of the model's fact-conflict hallucination correspond to an upward movement of the state in Figure \ref{fig:hallucination-categorize}. In essence, this symbolizes the enhancement of the model's capacity to accurately represent its internal knowledge state. A critical question, then, is how to discern the internal knowledge state of a model?

\section{Knowledge State Probing}
\label{sec:Knowledge State Probing}

This section delves into the complexities of discerning a model's internal knowledge state. It comprises two perspectives. The first, an external perspective, discuss how to determine if a model possesses specific knowledge based on the model generations; The second perspective, an internal view, questions if it is possible to determine whether a model possesses specific knowledge by its internal activation.

For the following pilot experiments, we have selected three families of models with different sizes: Llama2-chat\citep{touvron2023llama2} (13B and 7B); Qwen-chat\citep{bai2023qwen} (14B and 7B); Ziya-reader\citep{junqing2023never} (13B).

As for data, We randomly select passages from Chinese and English Wikipedia and instruct GPT3.5 to generate a knowledge-related question-answer pair. The answer generated by GPT3.5 based on the original Wikipedia is considered as the correct answer. We refer to the QA pairs obtained by this method as \textbf{wiki-QA} in this paper. Examples of instructions and corresponding output are shown in Appendix \ref{sec:Example of wiki-QA Instruction}.

\subsection{External perspective}

Determining whether a model has specific knowledge through its generation is a straightforward way. But it is challenging to accurately assess the model's knowledge state through a singular generation result, due to the uncertainty of generation caused by sampling \cite{lee2022factuality} and different generation tendencies \cite{chuang2023dola}. Multiple generation results can more faithfully reflect the knowledge state of the model.

In the presence of a correct answer, the consistency of the model's multiple generation with the correct answer is a reliable method for assessing knowledge state. The consistency of model generation with the correct answer can be computed using methods such as unigram overlap and cosine similarity of text representation.

However, the correct answer is hard to obtain in many scenarios, in which case self-consistency becomes a critical tool for assessing the validity of the generation. As evidenced by multiple research (\citet{manakul2023selfcheckgpt}, \citet{agrawal2023language}, \citet{hase2023methods}, \citet{elaraby2023halo}), there is a general conclusion that higher consistency across multiple generations is often indicative of validity of the generation. 
Intuitively, if the model has the corresponding knowledge, multiple generation are likely to contain consistent facts, resulting in higher consistency. Whereas, the contents of the hallucinations often varies, leading to lower self-consistency. 
We evaluate the self-consistency of a certain generation by the average of the cosine similarity representations among other generated answers.

\subsection{Internal perspective}
\label{sec:Internal perspective}

Previous work (\citet{azaria2023internal}, \citet{kadavath2022language}, \citet{li2023inference}) prove that LLMs can discern the factual accuracy of certain statements, even when the false statements are self-generated. This supports the existence of state 2 in Figure \ref{fig:hallucination-categorize} where the model has the corresponding knowledge but generates incorrect outputs. 
But are LLMs capable of discerning its own state of knowledge? The question can be rephrased as follows: for a given knowledge-related question, can a model discern its capability to output the correct answer before the actual generation of an answer? The following linear probing experiments on multiple models implies that the answer is yes.

We sample questions from the wiki-QA data, and LLM generates $k=5$ answers for each question separately. We use the consistency method described earlier to pre-label the questions. The sum of these normalized consistency scores computed to derive the final score. 

To categorize the questions, straightforward thresholds are utilized. The upper threshold is set at the 65th percentile score, and the lower at the 35th percentile score. Under this setup, responses with scores exceeding the upper threshold are labeled as correct, while those falling below the lower threshold are labeled as incorrect. 
If all of the k generated responses related to a specific question are deemed correct, the model is presumed to possess the relevant internal knowledge, and thus the question is labeled as 'Known'. Conversely, if all k responses are incorrect, the model is considered to lack the necessary internal knowledge, and hence the question is labeled as 'Unknown'.

A single linear layer classifier (probe) is trained on the internal representation corresponding to the last token of each question. Its task is to predict the corresponding Known/Unknown label.

For our experiments, we select three types of internal representations:

\textbf{The attention output}, which refers to the output of the dot product attention and before the attention linear layer in the decoder layer. This setup aligns with the probe's positioning within \citet{li2023inference}; \textbf{The MLP output}, i.e., the feed-forward layer's output within the decoder layer, occurring prior to residual linkage; \textbf{The hidden states}, defined as each decoder layer's output.

The results of the internal knowledge probe experiment are shown in Figure \ref{fig:probe-results}, which presents the accuracy of the trained probes across different models with different internal representation and at different layers. 
 
Comparative analysis of the experimental results across models of varying sizes yields consistent observations:

1. The linear probes of the internal state accurately predict the knowledge representation of the model. The probes' maximum accuracy surpasses 0.85 in most setups. This suggests that information about whether the model has the corresponding knowledge is linearly encoded in the internal representation of the model with high accuracy.

2. The accuracy of the probes increases rapidly in the early to middle layer, indicating that the model needs some layers of computation before it can determine its own knowledge states.

3. Hidden state probes exhibit the highest accuracy in discerning the knowledge state of the model, sustaining high accuracy from the middle layer to the output layer, which opens up the possibility of utilizing internal knowledge state when generating responses.


\subsection{DreamCatcher}

We integrated the above methods of knowledge state probing and consistency judgments to develop an automatic hallucination labeling tool, DreamCatcher.

We start by collect the LLM\'s generation for each question in the question set, in our case, the wiki-QA dataset. This process features two modes: normal generation and uncertainty generation. Normal generation is when the prompt contains only the question and model generates k responses, while uncertainty generation refers to where the prompt contains a request for the model to output answers that show uncertainty or lack of knowledge.

Subsequently, we assess the degree of hallucination of the generated responses using multiple scorers using the methods described above.
Concretely, we compute the following scores:

$$
\begin{array}{ll}
s_{s2g}   &= \text{avg}_{ij}(\text{cos}(\textbf{r}_{G_i}, \textbf{r}_{G_j})) \\
s_{p} &= \text{probe}(\textbf{r}_{Q}) \\
s_{o2a}   &= \text{count}(token_{overlap}) / \text{count}(token_A) \\
s_{s2a}   &= \text{cos}(\textbf{r}_G, \textbf{r}_A) \\
\end{array}
$$

where $Q$ denotes the question, $A$ the correct answer, $G$ the generation and $\textbf{r}$ the embedding representation of text.

\textbf{$s_{p}$ (Probe Score)}: rates the questions by utilizing the probes trained in Section \ref{sec:Internal perspective}, which are intended to discern the model's knowledge state for the corresponding questions.

\textbf{$s_{o2a}$ (Overlap with Answer Score)}: calculates the ratio of token overlap between the generated output and the correct answer ($A$).

\textbf{$s_{s2a}$ (Similarity to Answer Score)}: computes the cosine similarity between the embedding of the generated response ($G$) and the correct answer ($A$), using the bge-large model for text embedding.

The scores are normalized and summed to provide an overall factuality score for each generation. The generations are then classified as "correct" or "incorrect" based on whether their total score is above or below the median score, respectively. Questions are categorized as "Known", "Unknown", or "Mixed" based on whether the responses are consistently correct, incorrect, or a combination of correct and incorrect across multiple generations, with "Mixed" being a less frequent occurrence.

The categories correspond to three ranking hierarchies as shown in Figure \ref{fig:RLKF}: Known (corresponding to state 1 in Fig. \ref{fig:hallucination-categorize}): factual > uncertainty; Mixed (state 2): factual > uncertainty > hallucination; Unknown (state 4): uncertainty > hallucination. Here, "factual" refers to the generation with the highest factuality score, while "hallucination" denotes the generation with the lowest score. 

We randomly sampled 200 entries, half Chinese and half English, from the DreamCatcher labeled data. Then the human annotator annotate the same data, without access to the labels of DreamCatcher. The consistency between DreamCatcher and human annotator is shown in Table \ref{tab:Dreamcatcher-acc}, with an overall accuracy of 81\%. 

\begin{table}[htbp]
\resizebox{\linewidth}{!}{
\begin{tabular}{cccc}
\toprule
Language & Accuracy & Precision & Recall \\
\midrule
All      & 81\%     & 77\%      & 86\%   \\
Chinese       & 77\%     & 79\%      & 76\%   \\
English       & 86\%     & 76\%      & 98\%   \\
\bottomrule
\end{tabular}
}
\caption{The consistency between DreamCatcher and human annotator. For precision and recall, we treat "correct" as a positive label and "incorrect" as negative.}
\label{tab:Dreamcatcher-acc}
\end{table}

\section{Method}
\label{sec:Method}

\begin{figure*}[htbp]
\includegraphics[width=\linewidth]{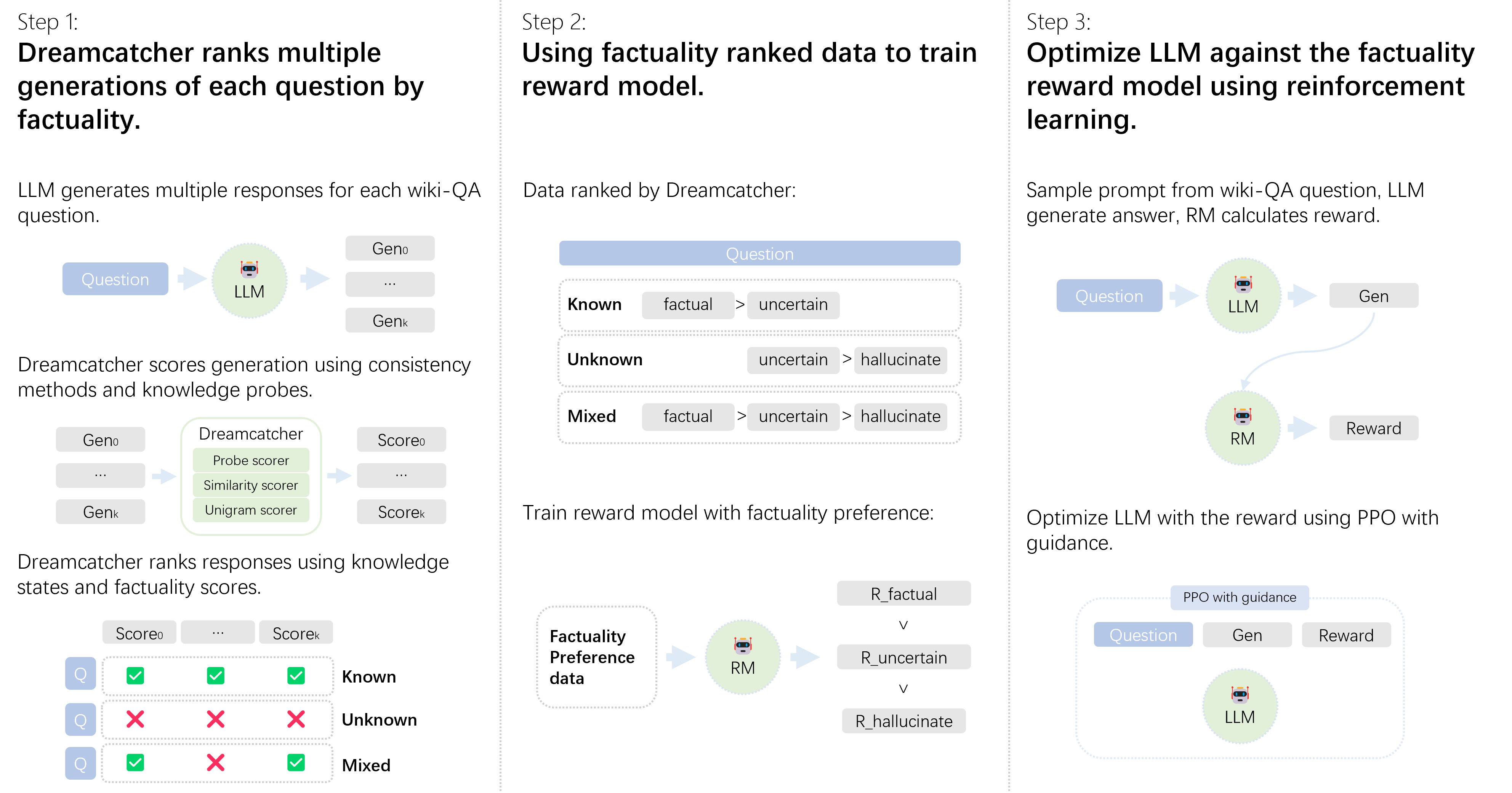} 
\caption{RLKF training}
\label{fig:RLKF}
\end{figure*}

From the above knowledge-probing experiments, we discover that LLMs are capable of evaluating their own knowledge states in response to specific knowledge-based questions. This implies that LLMs demonstrate a self-awareness of their knowledge state, which does not consistently translate into their generational output. 

Frequently, when faced with questions outside of internal knowledge, LLMs tends to generate hallucinations. Additionally, even with questions within internal knowledge, LLMs may potentially generate incorrect responses due to other influences. One possible explanation could be that LLMs did not learn to generate with respect to the internal knowledge state during model training. Instead, the fine-tuning process often requires the model to generate seemingly reasonable answers to all factual questions.

We therefore emphasize on enhancing the model's utilization of internal knowledge state so that the model can choose to rely on internal knowledge to answer or honestly express its lack of relevant knowledge.\footnote{This intuition could also be used for efficient RAG, enabling direct responses when the LLM possesses relevant internal knowledge, while relying on the retrieval tool in case of a knowledge gap.}

Consequently, we propose the RLKF (Reinforce Learning from Knowledge Feedback) training framework. This introduces model knowledge state assessments into the reinforcement learning feedback mechanism, enhancing model honesty and factuality. The RLKF training process shares similarities with the standard RLHF (Reinforce Learning from Human Feedback), and can integrate smoothly with the existing RLHF framework, but reduces data collection costs by substituting human labeling with automatic knowledge labeling.

The RLKF training framework consists of the following components, as shown in Figure \ref{fig:RLKF}.

\textbf{Knowledge state annotation}: We annotate factual preference data using the Dreamcatcher tool.

\textbf{Knowledge Feedback Modeling}: Having obtained the factual preference data, we train the reward model following \citealp{ouyang2022training}. The language modelling head in reward model is replaced with a linear layer to produce a scalar output, corresponding to the reward of the generated response. In line with \cite{kopf2023openassistant}, an additional regularization parameter is introduced to prevents the predicted values from diverging too much.

By initiating the PPO Policy training and the reward model training from the same model, we can ensure that the Reward model can leverage the same internal knowledge.

\textbf{PPO Optimizing}: Based on our factual reward model, we optimize the policy, i.e., the initial generative model, using the PPO algorithm once again following \citealp{ouyang2022training}.
To improve the efficiency of model exploration towards honesty, we use guidance technique in reinforcement learning. Concretely, we concatenate the first few tokens of the preferred responses to the input prompts in a portion of the training data. The added tokens do not participate in the loss calculation, but can guide the model to generate desired responses, thus improving learning efficiency. 

The core of the training framework is to establish the factual preference reward mechanism. The reinforcement learning algorithms in the RLKF framework can also be replaced by other optimization algorithms such as DPO \cite{rafailov2023direct}, reject sampling, etc. We choose PPO to be consistent with the common practice in RLHF training.

\section{Experiments}
\label{sec:Experiments}

In the following experiments, We chose three different models of varying sizes: llama2-chat (13B and 7B); Qwen-chat (14B and 7B); and Ziya-reader (13B), which is consistent with the choice of models for the knowledge-probing experiments detailed in Section \ref{sec:Knowledge State Probing}.

\begin{table}[htbp]
\resizebox{\linewidth}{!}{
\begin{tabular}{lccc}
\toprule 
\multicolumn{1}{c}{\textbf{Model}} & \textbf{Known} & \textbf{Unknown} & \textbf{Mixed} \\
\midrule
Qwen-chat-14B                      & 82.7\%          & 87.1\%            & 77.8\%          \\
Qwen-chat-7B                       & 65.7\%          & 81.6\%            & 61.1\%          \\
Llama2-chat-13B                    & 85.4\%          & 85.4\%            & 60.0\%          \\
Llama2-chat-7B                     & 78.9\%          & 89.2\%            & 57.6\%          \\
Ziya-reader-13B                    & 93.5\%          & 82.4\%            & 64.5\%         \\
\bottomrule
\end{tabular}
}
\caption{Accuracy of trained reward model for each knowledge state category.}
\label{tab:rm-acc}
\end{table}

\begin{table*}[htbp]
\resizebox{\textwidth}{!}{
\begin{tabular}{clcccccccc|c}
\toprule
\multicolumn{2}{c}{Models}                & MMLU           & WinoGrande     & ARC            & BBH            & GSM8K          & MATH           & C-Eval         & CMMLU          & Avg            \\
\midrule
\multirow{2}{*}{Qwen-chat-14B}   & before & 64.2\%          & 53.8\%          & 76.5\%          & 34.5\%          & 47.3\%          & 18.9\%          & 65.0\%          & 64.1\%          & 53.0\%          \\
                                 & after  & \textbf{64.5\%} & \textbf{59.1\%} & \textbf{87.2\%} & \textbf{37.3\%} & \textbf{49.9\%} & \textbf{20.3\%} & 64.6\%          & \textbf{66.4\%} & \textbf{56.2\%} \\
\multirow{2}{*}{Qwen-chat-7B}    & before & 54.2\%          & 49.6\%          & 63.1\%          & 28.8\%          & 50.0\%          & 12.6\%          & 57.8\%          & 58.1\%          & 46.8\%          \\
                                 & after  & \textbf{55.3\%} & \textbf{52.2\%} & \textbf{75.4\%} & 28.1\%          & \textbf{50.9\%} & 12.5\%          & 57.5\%          & 56.0\%          & \textbf{48.5\%} \\
\multirow{2}{*}{Llama2-chat-13B} & before & 52.3\%          & 51.9\%          & 72.4\%          & 21.7\%          & 35.2\%          & 3.2\%           & 34.6\%          & 34.5\%          & 38.2\%          \\
                                 & after  & \textbf{52.8\%} & \textbf{54.3\%} & 72.1\%          & \textbf{23.4\%} & \textbf{35.6\%} & 3.1\%           & 34.3\%          & \textbf{34.6\%} & \textbf{38.8\%} \\
\multirow{2}{*}{Llama2-chat-7B}  & before & 45.9\%          & 51.5\%          & 59.2\%          & 23.3\%          & 25.9\%          & 1.6\%           & 32.1\%          & 31.6\%          & 33.9\%          \\
                                 & after  & \textbf{46.2\%} & \textbf{52.4\%} & \textbf{61.1\%} & \textbf{24.4\%} & 23.7\%          & \textbf{2.0\%}  & \textbf{34.0\%} & \textbf{32.1\%} & \textbf{34.5\%} \\
\multirow{2}{*}{Ziya-reader-13B} & before & 49.5\%          & 50.8\%          & 64.7\%          & 44.7\%          & 29.3\%          & 4.3\%           & 44.7\%          & 46.1\%          & 41.7\%          \\
                                 & after  & \textbf{50.3\%} & \textbf{51.9\%} & \textbf{67.9\%} & 42.6\%          & \textbf{33.2\%} & 3.8\%           & 42.6\%          & 45.1\%          & \textbf{42.2\%} \\
\bottomrule
\end{tabular}
}
\caption{\label{benchmarks}Evaluation of RLKF-trained models on various knowledge and reasoning related tasks: MMLU \cite{hendrycks2020measuring}, WinoGrande \cite{sakaguchi2021winogrande}, ARC \cite{chollet2019measure}, BBH \cite{suzgun2022challenging}, GSM8K \cite{cobbe2021training}, MATH \cite{hendrycks2021measuring}, C-Eval \cite{huang2023c}, CMMLU\cite{li2023cmmlu}. Tasks are evaluated by the open-source evaluation tool TLEM \cite{tlem}, employing a 0-shot setting with greedy generation.}
\end{table*}

\subsection{Data collection}

We used the wiki-QA data collection method same as in Section \ref{sec:Knowledge State Probing}, obtaining about 7,000 QA pairs each for Chinese and English. To add variety to the questions, we have also modified the prompt to include multiple choice question types. Since our approach relies on the internal knowledge of the models and the boundaries of the internal knowledge are different for each model, we need to perform automatic annotation for each model individually. The generated responses are labeled using Dreamcatcher to obtain factual preference data. The statistics of the factual preference data are shown in Table \ref{tab:factual-preference-data}.

\subsection{RLKF Training}

We train the reward model using the factual preference data in Table \ref{tab:factual-preference-data}. To maintain the generalization of the RM, we include same amount of general purpose data as the wiki-QA data in the training. Accuracy of the trained RM on factual preference data testset are shown in Table \ref{tab:rm-acc}. Interestingly, the reward model is able to quickly achieve high accuracy for both known/unknown categories during training, suggesting that reward model may utilize the internal knowledge state of the initial model to determine whether the uncertainty response should be preferred.

Using the trained reward model, the RL process optimizes policy model using the PPO algorithm, where policy model is initialized from the same base model as reward model. The detailed training settings and hyper-parameters are described in \ref{sec:RLKF Training details}.


We conduct an evaluation of the trained model, focusing on its factuality and truthfulness. A comparative analysis of the models is performed between pre- and post- RLKF training on various tasks related to knowledge and reasoning as shown in Table \ref{benchmarks}. The RLKF-trained models demonstrate improvements on the majority of the benchmarks. While RLHF typically results in a reduction of benchmark performance, termed as 'alignment tax' \citep{askell2021general}, RLKF avoids this decline specifically on knowledge-related tasks, and even lead to improvements. Note that our training methodology does not employ any benchmark data, and the overall volume of training data utilized is small.

Regarding the truthfulness of trained models, we evaluated their performance using the widely recognized TruthfulQA task. Notably, all models, with the exception of llama2-chat-13B, show increase in honesty, as shown in Table \ref{TruthfulQA}.


\begin{table}[]
\begin{tabular}{clc}
\toprule
\multicolumn{2}{c}{Models}                & TruthfulQA \\
\midrule
\multirow{2}{*}{Qwen-chat-14B}   & before & 43.7\%         \\
                                 & after  & \textbf{49.1\%} \\
\multirow{2}{*}{Qwen-chat-7B}    & before & 49.1\%          \\
                                 & after  & \textbf{50.3\%} \\
\multirow{2}{*}{Llama2-chat-13B} & before & 21.5\%          \\
                                 & after  & 20.9\%          \\
\multirow{2}{*}{Llama2-chat-7B}  & before & 27.5\%          \\
                                 & after  & \textbf{28.3\%} \\
\multirow{2}{*}{Ziya-reader-13B} & before & 34.8\%          \\
                                 & after  & \textbf{37.9\%} \\
\bottomrule
\end{tabular}
\caption{\label{TruthfulQA}Evaluation of RLKF-trained models on TruthfulQA, again using TLEM \cite{tlem}, employing a 0-shot setting with greedy generation.}
\end{table}

\section{Related Work}
\label{sec:Related works}

Hallucination in large language models (LLMs) has been the focal point of research, spanning its causes, detection, and mitigation. Our work relates to all three aspects.

\textbf{Causes of hallucination}: 
Studies have linked LLM hallucination to various causes. \citet{mckenna2023sources} ascribes it to memorization of training data, indicating a direct correlation between the training data and the resultant hallucination.
Other works such as \citet{schulmanreinforcement} pinpoint improper model fine-tuning as contributive, and \citet{perez2022discovering} argues that RLHF induce model "sycophancy" which in turn degrades honesty.

Other studies link hallucinations to the generation process. For example, \citet{lee2022factuality} suggests that sampling-induced randomness could be responsible. One perspective provided by \citet{chuang2023dola} proposes that "lower-level" prior layer information might overshadow factual information from subsequent layers. Furthermore, some works relate hallucinations to the overconfidence of LLMs \cite{ren2023investigating}.

\textbf{Hallucination detecting}: 
In terms of detecting hallucination, the consistency of multiple generations has been recognized as an effective indicator. SelfCheckGPT \cite{manakul2023selfcheckgpt} capitalizes on the consistent nature of internal knowledge-based generations compared to the variable nature of hallucination, leading to the proposal of several consistency checks to identify hallucinations. 
The idea is echoed by \citet{agrawal2023language}, who suggest evaluating the generation consistency of generated references to spot hallucination. Similarly, \citet{elaraby2023halo} proposes a metric involving the calculation of sentence-level entailment between response pairs as a measure of hallucination.

Employing large language models (LLMs) to recognize their own hallucinations has been suggested in \citet{saunders2022self}, suggesting that discrimination is more accurate than generation for LLMs (G-D gap). This notion is furthered by \citet{kadavath2022language} and \citet{agrawal2023language} by directly prompting LLMs to assess the validity of their own output.

Another approach examines the factualness of statements by analyzing the model's internal representation. 
Studies \citet{li2023inference} and \citet{burns2022discovering} identify a "factualness" direction in the model's internal representation, with \citet{li2023inference} showcasing a high accuracy attention head through linear probing, and \citet{burns2022discovering} locating factualness direction through consistency of facts. Additionally, \citet{kadavath2022language} trains the model to predict the probability that it knows. Base on these works, we shifts focus onto the model's self-evaluation of knowledge state.


\textbf{Hallucination mitigation}:
The common approach of hallucination mitigation involves enhancing the model with additional information. \citet{elaraby2023halo} propose the use of larger models to provide additional information when hallucinations is detected.

Some research efforts focus on the optimization of decoding strategies to address hallucinations. \citet{chuang2023dola} suggests that contrastive decoding can augment the factualness of model generation. Others leverage the direction of factualness in the model representation space; \citet{li2023inference} enhances factualness by adjusting the output of attention heads along the direction of factualness during inference. Our work seeks to optimizes the utilization of the model's internal knowledge state, in line with the direction proposed by \citet{schulmanreinforcement} leveraging reinforcement learning to tackle hallucinations.

\section{Conclusion}
\label{sec:Conclusion}

In our research, we have thoroughly explored the capability of large language models (LLMs) to discern and express their internal knowledge, a key factor in mitigating factual hallucinations and ensuring reliable applications. Our research, manifested through a series of knowledge probing experiments, identifies the model's self-awareness of its knowledge state. We released the open-source tool DreamCatcher which scores and labels the degree of hallucination in the LLM's response to knowledge-oriented question and rank responses based on their factuality.

We further validated our findings through the introduction of a training framework: Reinforcement Learning from Knowledge Feedback (RLKF). Utilizing DreamCatcher to annotate factual preference data, we trained a reward model and leveraging reinforcement learning to enhances LLM's factuality and truthfulness. 
Our results indicate RLKF's effectiveness in improving the model's utilization of its internal knowledge state, enhancing its performance in various knowledge and honesty related tasks. We posit that RLKF is a promising solution to address LLM's hallucination issues and, combined with RLHF, offers significant potential for enhancing the model's overall capabilities.




\newpage

\bibliography{custom}

\newpage

\appendix
\onecolumn
\section{Appendix}
\label{sec:appendix}

\subsection{Example of wiki-QA Instruction}
\label{sec:Example of wiki-QA Instruction}

\begin{table*}[htbp]
\resizebox{\textwidth}{!}{
\begin{tabular}{p{12cm}}
\toprule
\textbf{Instruction template}:

Based on the following Wikipedia article snippet, ask a knowledge-based question and provide a corresponding answer.

Article snippet: 

\{Wikipedia passage\}

Requirements:

1. there is a unique correct answer to the question, and the answer can be found in the given article fragment.

2. the question can be answered independently of the article fragment, i.e. the answer to the question cannot depend on contextual information, e.g. a question about a character in a literature needs to specify the work to which the character belongs, and a question such as "What is the article about?" cannot be asked. 

3. Provide the question, answer, and category (e.g., literature, physics, etc.) at the same time, and reply in the following format: \{"question":question,"answer":answer,"type":category\}.

If you are unable to ask a question that meets the above requirements, you can simply reply "Unable to ask".

Reply:

\textbf{Wikipedia passage}:

House Arrest (1996 film) 
House Arrest is a 1996 American comedy film directed by Harry Winer, written by Michael Hitchcock, and starring Jamie Lee Curtis,  Kevin Pollak, Jennifer Tilly, Christopher McDonald, Wallace Shawn, and Ray Walston with supporting roles done by Kyle Howard, Amy Sakasitz, Mooky Arizona, Russel Harper, and an up-and-coming Jennifer Love Hewitt. It tells the story of two children who trap their parents in their basement upon their plans for a separation as the other children they know get involved by trapping their respective problem parents as well.
The film was released on August 14, 1996 and went on to gross just over \$7 million at the box office. The film was panned by critics.
The film was shot at various locations in the U.S. states of California and Ohio. Monrovia, California was the location for several exterior house scenes while most interior shots were done at the CBS/Radford lot in Studio City, California. The story was set in Defiance, Ohio, although another town, Chagrin Falls, Ohio, actually doubled for it.

\textbf{GPT3.5 response}:

\{"question":"Who directed the film House Arrest?","answer":"Harry Winer","type":"film"\} \\
\bottomrule
\end{tabular}
}
\caption{Example of instruction and corresponding GPT3.5 output of English wiki-QA.}
\end{table*}

\begin{CJK*}{UTF8}{gbsn}

\begin{table*}[htbp]
\resizebox{\textwidth}{!}{
\begin{tabular}{p{12cm}}
\toprule

\textbf{Instruction template}:

根据下面的维基百科文章片段，提出一个简短的知识型问题并给出对应回答，要求这个问题存在唯一正确答案，并且答案可以在给出的文章片段中找到。

文章片段：

\{Wikipedia passage\}

问题需要在脱离文章片段的情况下仍能够被回答，例如针对文学作品中人物提问需要指明所属的作品，以免引起歧义。问题的回答不能依赖于上下文的信息，不能提出类似“这篇文章的内容是什么？”的问题。

同时给出问题，回答和问题分类（比如文学类或物理类等），按如下格式回复：\{"question":问题,"answer":回答,"type":分类\}。如果无法提出满足上述要求的问题，可以直接回复“无法提问”。

回复：

\textbf{Wikipedia passage}:

M25 

M25，也称为IC 4725，是一个由恒星组成，在南天人马座的疏散星团。Philippe Loys de Chéseaux在1745年对这个星团进行了第一次有记录的观测，查尔斯·梅西耶1764年将它收录进他的星云天体清单[6]。这个星团位于模糊的特征附近，因此有一条暗带通过中心附近[3]。

M25距离地球大约2,000光年，年龄约为6,760万岁[2]。这个星团在空间的维度大约是13光年，估计质量是1,937 M☉，其中大约24\%是星际物质[4]。星团成员中的人马座U是一颗分类为造父变星的变星[7]，还有两颗红巨星，且其中一颗是联星系统[8]。

\textbf{GPT3.5 response}:

\{"question":"M25是位于哪个星座的疏散星团？","answer":"南天人马座","type":"天文学"\} \\

\bottomrule
\end{tabular}
}
\caption{Example of instruction and corresponding GPT3.5 output of Chinese wiki-QA.}
\end{table*}
\end{CJK*}

\subsection{More probing results}
\label{sec:More probing results}

\begin{figure*}[htbp]
\begin{subfigure}{0.45\textwidth}
\includegraphics[width=0.95\linewidth]{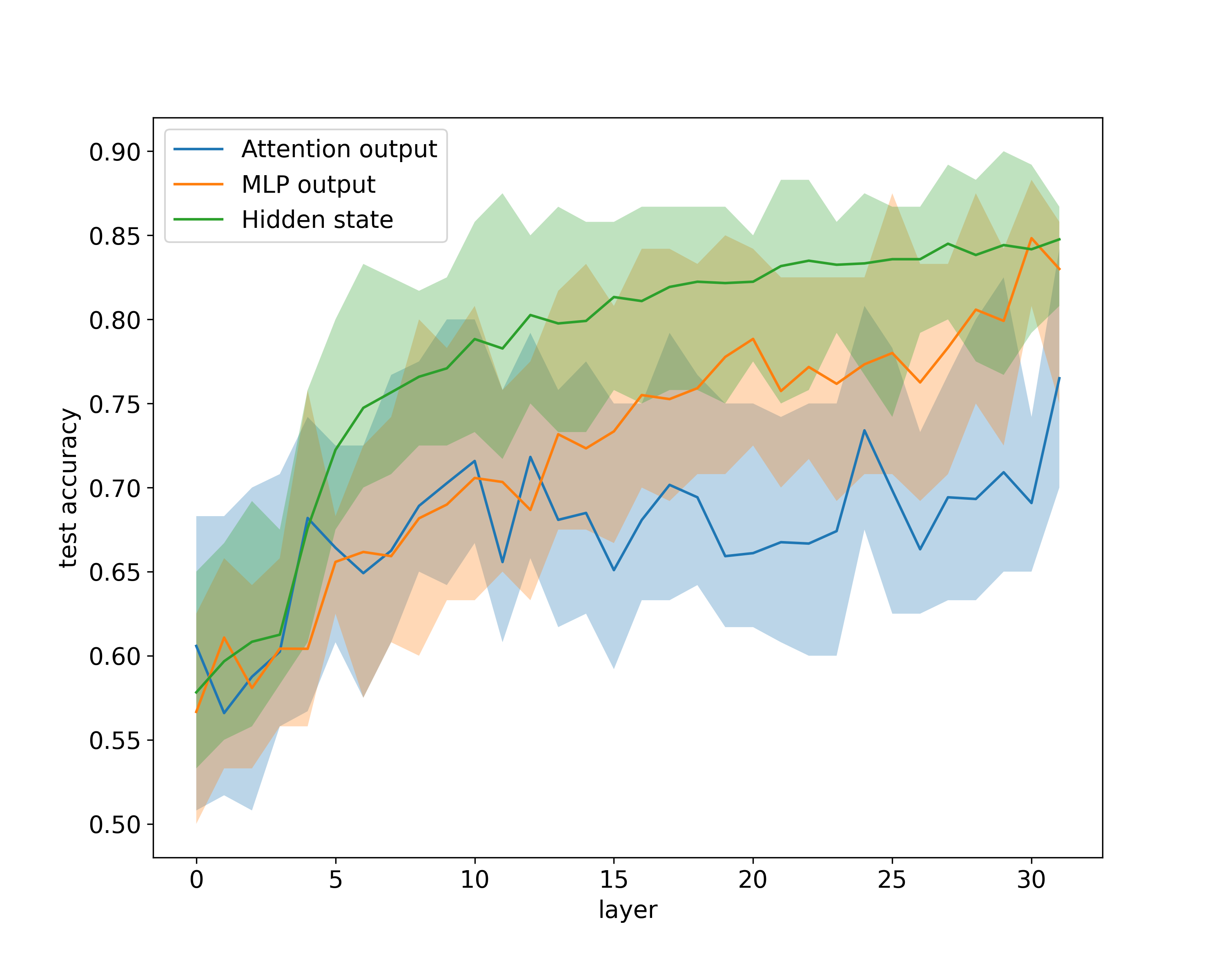} 
\caption{Llama2-chat-7B}
\label{fig:subim1}
\end{subfigure}
\begin{subfigure}{0.45\textwidth}
\includegraphics[width=0.95\linewidth]{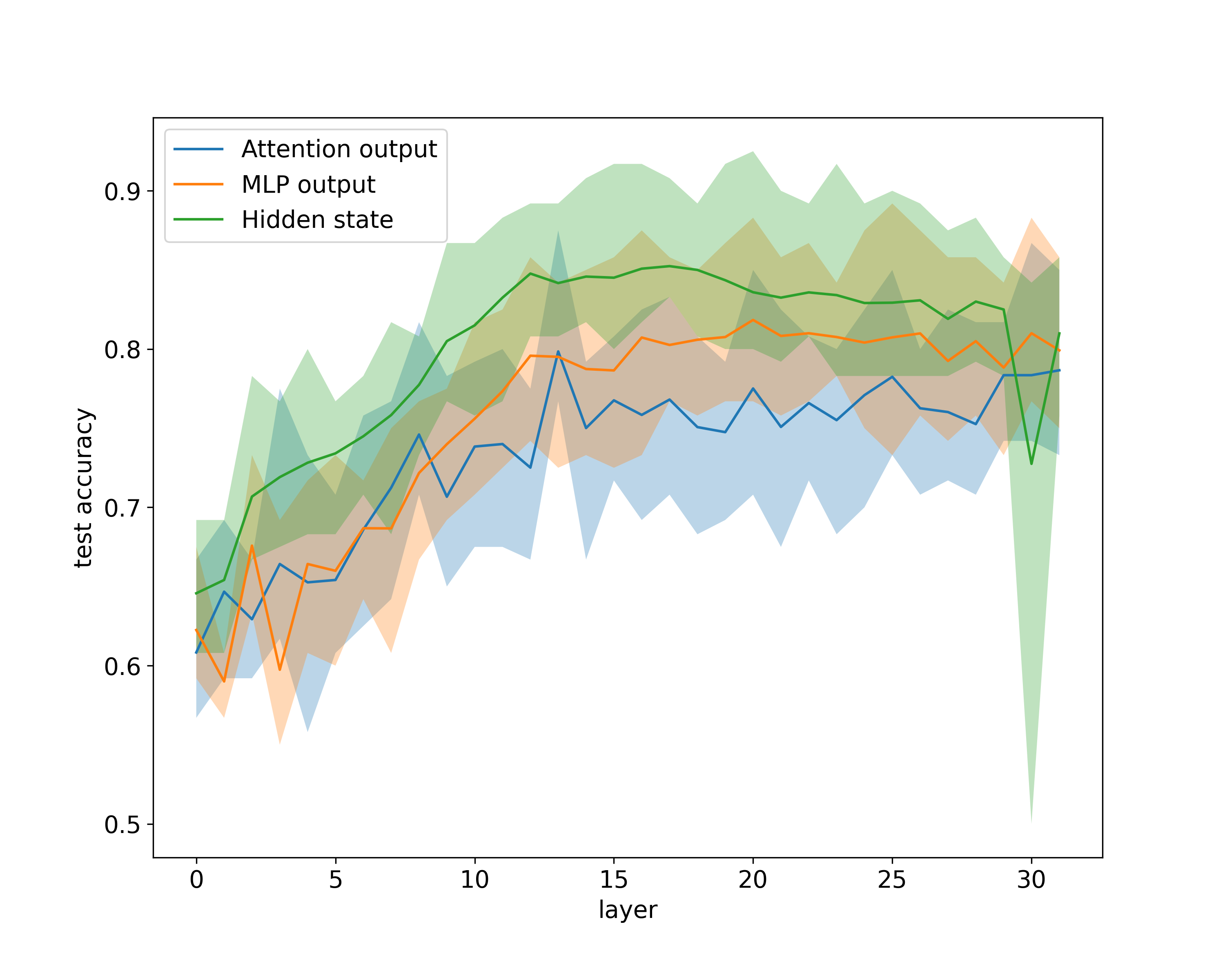}
\caption{Qwen-chat-7B}
\label{fig:subim2}
\end{subfigure}
\caption{Accuracy of knowledge state probing in 7B models. The light-colored area in the figure shows the range of accuracy for ten repetitions of the experiment, and the solid line shows the mean accuracy. }
\label{fig:probe-results-7B}
\end{figure*}


\subsection{Statistics of factual preference data}
\label{sec:Statistics of factual preference data}
\begin{table}[htbp]
\resizebox{\linewidth}{!}{
\begin{tabular}{lcccc}
\toprule 
\multicolumn{1}{c}{\textbf{Model}} & \textbf{Total} & \textbf{Known} & \textbf{Unknown} & \textbf{Mixed} \\
\midrule
Qwen-chat-14B                      & 12799          & 49\%           & 43\%             & 8\%            \\
Qwen-chat-7B                       & 7201           & 52\%           & 40\%             & 8\%            \\
Llama2-chat-13B                    & 6600           & 48\%           & 44\%             & 8\%            \\
Llama2-chat-7B                     & 6680           & 45\%           & 45\%             & 10\%           \\
Ziya-reader-13B                    & 12558          & 49\%           & 41\%             & 10\%          \\
\bottomrule
\end{tabular}
}
\caption{Statistics of factual preference data and percentage of each knowledge state category used for reward modeling. The Llama2 models use English-only wiki-QA data, Qwen-chat-7B uses Chinese-only data, and Qwen-chat-14B and Ziya-reader-13B use a mixture of English and Chinese data.}
\label{tab:factual-preference-data}
\end{table}

\subsection{RLKF Training details}
\label{sec:RLKF Training details}
We use the AdamW optimizer, with $\beta_1 = 0.9$, $\beta_2 = 0.99$, $eps$ = $1e-5$ for all models. The learning rate for reward model training is $5e-6$ with 1\% warmup and linear decay scheduler. The batch size is 16 for 13/14B models and 64 for 7B models. We train the reward model for 1 epoch. For PPO training, we use learning rate of $1e-6$ with cosine scheduler. The batch size is 32 for 13/14B models and 64 for 7B models. We set the KL penalty to 0 for all models.

\subsection{More Observation}

We observe that, some of the responses to the unknown questions are indicating uncertainty in RLHF-trained models, but there is also a significant percentage of responses that are hallucinations. This indicates an increase in model honesty achieved through RLHF, but there is still room for improvement.

\end{document}